\documentclass[letterpaper, 10 pt, conference]{ieeeconf}  

\IEEEoverridecommandlockouts                              

\usepackage{amsfonts}
\usepackage{graphicx}
\usepackage{cite}
\usepackage{rotating}
\usepackage{makecell}
\usepackage{svg}
\usepackage{amssymb}
\usepackage{pifont}
\newcommand{\cmark}{\ding{51}}%
\newcommand{\cbmark}{\ding{52}}%
\newcommand{\xmark}{\ding{53}}%
\usepackage{array}
\usepackage{hyperref}

\usepackage{pgfplots}
\usepgfplotslibrary{external}
\usepackage{subcaption}
\usepackage[export]{adjustbox}

\title{\LARGE \bf
From Rolling Over to Walking: Enabling Humanoid Robots to Develop Complex Motor Skills
}

\author{Fanxing Meng$^{1}$ and Jing Xiao$^{1}$
\thanks{$^{1}$Robotics Engineering Department,
        Worcester Polytechnic Institute, Worcester, MA, USA
        {\tt\small \{fmeng, jxiao2\}@wpi.edu}}%
}

\begin{document}
\bstctlcite{MyBSTcontrol}

\maketitle
\thispagestyle{empty}
\pagestyle{empty}

\begin{abstract}

This paper presents an innovative method for humanoid robots to acquire a comprehensive set of motor skills through reinforcement learning. The approach utilizes an achievement-triggered multi-path reward function rooted in developmental robotics principles, facilitating the robot to learn gross motor skills typically mastered by human infants within a single training phase. The proposed method outperforms standard reinforcement learning techniques in success rates and learning speed within a simulation environment. By leveraging the principles of self-discovery and exploration integral to infant learning, this method holds the potential to significantly advance humanoid robot motor skill acquisition.\footnote{Video available at \url{https://youtu.be/d0RqrW1EzjQ}}
\end{abstract}


\section{INTRODUCTION}

Humanoid robots, characterized by their adaptability to indoor environments and capacity to utilize tools designed for humans, present vast potential for home use. However, training these robots to execute actions that a human infant can perform remains a significant challenge. Existing reinforcement learning approaches have demonstrated potential in training robots in locomotion and manipulation skills, yet these often necessitate defining reward functions or providing human demonstrations for each task.

Groundbreaking research in the field of developmental robotics suggests that robots can learn about the physical world in a manner analogous to human infants\cite{SensorimotorContingencies}. This approach opens avenues for robots to acquire a diverse range of motor skills without the need for explicit programming or human demonstration for each task. Key research includes child-sized humanoid robots that emulate human developmental stages by amalgamating knowledge-based and competence-based intrinsic motivation learning signals\cite{intrinsic_motivation}.

In this paper, we propose an achievement-triggered multi-path reward function for reinforcement learning, which is capable of managing a broad range of motor skills based on developmental learning principles. This approach enables a humanoid robot's learning of rolling over, kneeling, crawling, standing up, and walking, eliminating the need for a complicated training pipeline or human demonstration data. We validate the efficacy of our approach by training a humanoid robot through a progressively complex task curriculum within a simulation environment.

\section{Related Work}

This section surveys the relevant literature, providing context and background for our research.

\subsection{Developmental Robotics}

Developmental robotics is focused on applying learning principles that enable humans to autonomously acquire complex motor skills to robots. These principles, derived from the way human infants interact with the world, encompass the ego-centric to object-centric transfer for motion representation\cite{egocentric}, the impact of morphological changes on action modeling\cite{dynamic_system}, the importance of embodiment for learning to roll over and crawl\cite{embody_emerge}, the necessity of the central pattern generator signal for crawling\cite{CPG}, and the level of an agent's existing skills serving as a determinant of its learning effectiveness\cite{Piaget}. Although these principles have been successfully applied to simpler tasks using less complex robotic hardware, their broad adoption remains limited in the reinforcement learning domain and application to sophisticated robots to perform intricate motions.

Our method integrates these principles within a reinforcement learning framework, enabling the successful training of advanced humanoid robots to execute a series of complex tasks, thus distinguishing our work from existing solutions.

\subsection{Curriculum Learning}

Curriculum learning, a closely related concept in machine learning, addresses challenging learning problems by progressively increasing data sample complexity to reduce inefficient exploration\cite{curriculum}. In robotics, curriculum learning has been frequently employed to solve manipulation problems. Various generalized methods like curriculum goal masking\cite{CGM} and adaptive hierarchical curriculum\cite{MultiPhase}, either omit or prioritize different components of the ultimate goal in various training phases. Other methods provide the learner with a linear sequence\cite{AbstractAdaptive} or transition dynamics graph\cite{PlanHierarchicallyFromCurriculum} of sub-tasks. Besides generating intermediate goals\cite{HGG}, curricula can be designed to start near the goal state\cite{VaPRL}\cite{SGC} or with the assistance of macro actions\cite{MacroActions}. Curricula can also be defined by escalating the number of controllable joints\cite{CASSL}, manipulable objects\cite{stacking}, and cooperative agents\cite{CooperativeTransportation}. Automatic curriculum learning methods hold promise for solving multi-agent\cite{curiosity_acl} and multi-objective\cite{hypernet} tasks.

For locomotion problems, a hierarchical curriculum with modular fractal networks has facilitated quadrupeds to learn walking and turning\cite{ContinualHierarchical}. Terrain complexity\cite{SixLegCL} or assistive force\cite{AssistiveForce} can also be used to structure the curriculum.

Our method employs a static, human-prior based curriculum that incorporates multiple stages of human infant motor skill development within every episode, thus eliminating the need for training multiple networks or adjusting the curriculum based on the agent's performance.

\subsection{Reinforcement Learning}

Recent advancements in deep reinforcement learning have empowered robots to acquire complex motor skills like locomotion\cite{daydreamer} and manipulation\cite{Bi-DexHands}. Despite these accomplishments, humanoid robot development remains restricted due to the instability of bipedal locomotion and higher degrees of freedom. Initial reinforcement learning applications to humanoids typically assumed a simplified model that reduced degrees of freedom and short horizon motions\cite{HumanoidRobotControlRL}. These efforts also included task decomposition by sub-goals for leg standing\cite{StandUpHierarchical}.

Subsequent studies applying Q-learning aimed to tackle the high-degrees-of-freedom challenge by constraining to planar motion \cite{PowerUsageReduction} or leveraging state space symmetry\cite{BilateralSymmetry}. Recently, policy-based learning methods like Proximal Policy Optimization\cite{PPO} and Generalized Advantage Estimation\cite{GeneralizedAdvantageEstimation} have become standard algorithms for high-dimensional motion learning. However, traditional benchmarks typically commence the learning process with the humanoid starting in a standing posture, potentially imposing strong preconditions. Furthermore, most implementations used a simplified capsule-like geometry and unrealistic actuator features\cite{DistributedReinforcementLearning}.

Historically, programming separate standing up behaviors for supine and prone positions was necessary. This imposed significant constraints on the geometric design and actuator performance, especially for the supine routine, which required the robot's arms to lift the body from the back to the squatting pose\cite{ReliableStandingUpRoutines}. These joint trajectories continue to be used as reinforcement learning objectives today\cite{FastReliableStandUp}. Aside from handcrafted reference motion, techniques like kinesthetic teaching by a human operator\cite{KinestheticBootstrapping} and human motion capture data have been used for correspondence learning\cite{MultimodalRewardFunction} or to bootstrap multi-stage learning algorithms\cite{MultiStageStandUp}. Most recently, Adversarial Motion Priors\cite{AMP}\cite{MultipleAMP} have emerged as the preferred method for learning complex and natural motion from unstructured human demonstrations.

State-of-the-art algorithms like SAC-X\cite{SACX} aim to solve multi-skill learning in sparse reward problems by employing a learned scheduler that switches between tasks. Research by \cite{LeggedRobotFallRecovery} applied a contact transition graph to model fall recovery policies, but the initial states for training required configurations near the goal state. \cite{AgileSoccerSkills} achieved learning sophisticated soccer skills including fall recovery, walking, and kicking. However, the get-up behavior was trained using pre-programmed pose targets, and then merged with a separately trained soccer policy for dynamic movement skills to emerge.

Current methodologies demonstrate the potential for learning a diverse range of long-horizon motion skills, albeit with certain limitations. In Table \ref{tab:comparison}, various algorithms and robot models employed by previous studies are listed and examined based on their prerequisites and the skills acquired. A prevalent trend observed is that mastering skills of greater complexity often requires human demonstrations or Motion Capture (MoCap) data, whereas more basic skills benefit from the motion produced by other controllers or predefined joint trajectories. Without preprocessed data, learning has only been successful on robots with fewer degrees of freedom or for simpler skills on higher-DoF robots.

\begin{table*}
\centering
    {
    \setlength\extrarowheight{5pt}
     \smallskip
     \smallskip
    \begin{tabular}{|l|l|l|l|l|l|l|l|l|l|l|l|l|}
        \hline
        {\scriptsize Work} & {\scriptsize Algorithm} & {\scriptsize Robot model} & {\scriptsize Sim} & {\scriptsize Real} & {\scriptsize Initial pose} & {\scriptsize Target pose} & {\scriptsize Natural} & {\scriptsize Preprocessing} & \makecell[l]{{\scriptsize Training time} \\ {\tiny steps / epochs / episodes}}  \\
        \hline
        \cite{HumanoidRobotControlRL} & TD & \textbf{3D}, 20 DoF, \textbf{realistic} & ODE & \cbmark & \includesvg[height=10pt]{emoji/sitstool} & \includesvg[height=10pt]{emoji/stands} & \xmark & \textbf{None} & / / 2922  \\
        \hline
        \cite{StandUpHierarchical} & Q-learning & 2D, 2 DoF, unrealistic & \cmark & \cbmark & \includesvg[height=5pt]{emoji/legf} & \includesvg[height=8pt]{emoji/leg} & \xmark & \textbf{None} & / / 750 \\
        \hline
        \cite{PowerUsageReduction} & Q-learning & 2D, 3 DoF, unrealistic & \cmark & \cbmark & \includesvg[height=8pt]{emoji/squat} & \includesvg[height=10pt]{emoji/stands} & \xmark & \textbf{None} & / / \\
        \hline
        \cite{BilateralSymmetry} & Q-learning & \textbf{3D}, 20 DoF, \textbf{realistic} & \cmark & \cbmark & \includesvg[height=4pt]{emoji/supine}\;\includesvg[height=5pt]{emoji/pronea} & \includesvg[height=10pt]{emoji/stands} & \xmark & \textbf{None} & / / 850 \\
        \hline
        \cite{PPO} & PPO & \textbf{3D}, 24 DoF, unrealistic & Bullet & \xmark & \includesvg[height=10pt]{emoji/stands} & \includesvg[height=7pt]{emoji/getup}\,\includesvg[height=10pt]{emoji/run} & \xmark & \textbf{None} & 1e8 / /  \\
        \hline
        \cite{GeneralizedAdvantageEstimation} & GAE & \textbf{3D}, 10 DoF, unrealistic & MuJoCo & \xmark & \includesvg[height=4pt]{emoji/supine} & \includesvg[height=6pt]{emoji/sitground}\,\includesvg[height=10pt]{emoji/stands} \includesvg[height=10pt]{emoji/walk} & \xmark & \textbf{None} & / / 1000 \\
        \hline
        \cite{DistributedReinforcementLearning} & PPO & \textbf{3D}, 28 DoF, unrealistic & Flex & \xmark & \includesvg[height=10pt]{emoji/stands} & \includesvg[height=7pt]{emoji/getup}\,\includesvg[height=10pt]{emoji/run} & \xmark & \textbf{None} & 1e10 / / \\
        \hline
        \cite{ReliableStandingUpRoutines} & PD control & \textbf{3D}, 19 DoF, \textbf{realistic} & ODE & \cbmark & \includesvg[height=4pt]{emoji/supine}\;\includesvg[height=5pt]{emoji/pronea} & \includesvg[height=3pt]{emoji/pushuplow}\,\includesvg[height=6pt]{emoji/sitground}\,\includesvg[height=8pt]{emoji/squat}\,\includesvg[height=10pt]{emoji/stands} & \xmark & Target traj. & / / \\
        \hline
        \cite{FastReliableStandUp} & BioIK, PD & \textbf{3D}, 20 DoF, \textbf{realistic} & PyBullet & \cbmark & \includesvg[height=4pt]{emoji/supine}\;\includesvg[height=5pt]{emoji/pronea} & \includesvg[height=3pt]{emoji/pushuplow}\,\includesvg[height=6pt]{emoji/layback}\,\includesvg[height=8pt]{emoji/squat}\,\includesvg[height=10pt]{emoji/stands} & \xmark & Target traj. & / / \\
        \hline
        \cite{KinestheticBootstrapping} & Genetic & \textbf{3D}, 18 DoF, \textbf{realistic} & ODE & \cbmark & \includesvg[height=5pt]{emoji/pronea} & \includesvg[height=3pt]{emoji/pushuplow}\,\includesvg[height=6pt]{emoji/bend}\,\includesvg[height=10pt]{emoji/walk} & \xmark & Kinesthetic teach & / / \\
        \hline
        \cite{MultimodalRewardFunction} & Evolution & \textbf{3D}, 28 DoF, \textbf{realistic} & \xmark & \cbmark & \includesvg[height=10pt]{emoji/sitstool} & \includesvg[height=10pt]{emoji/stands} & \cbmark & Human MoCap & / / \\
        \hline
        \cite{MultiStageStandUp} & DOE, SGD & \textbf{3D}, 20 DoF, \textbf{realistic} & Webots & \cbmark & \includesvg[height=5pt]{emoji/pronea} & \includesvg[height=10pt]{emoji/stands} & \cbmark & Human MoCap & / / 84 \\
        \hline
        \cite{AMP} & GAIL, PPO & \textbf{3D}, \textbf{34} DoF, unrealistic & Bullet & \xmark & \includesvg[height=5pt]{emoji/pronea}\;\includesvg[height=6pt]{emoji/sitground}\;\includesvg[height=10pt]{emoji/stands} & \includesvg[height=8pt]{emoji/squat}\,\includesvg[height=10pt]{emoji/stands}\,\includesvg[height=10pt]{emoji/walk}\,\includesvg[height=9pt]{emoji/dribble}\,\includesvg[height=6pt]{emoji/leap}\,\includesvg[height=10pt]{emoji/cartwheel} & \cbmark & Motion data & 3e8 / /  \\
        \hline
        \cite{MultipleAMP} & PPO & \textbf{3D}, 16 DoF, \textbf{realistic} & PhysX & \cbmark & \includesvg[height=5pt]{emoji/pronea} & \includesvg[height=5pt]{emoji/duck}\,\includesvg[height=8pt]{emoji/squat}\,\includesvg[height=10pt]{emoji/stands} & \cbmark & RL/MPC traj. &  / 1000 /  \\
        \hline
        \cite{SACX} & SAC & \textbf{3D}, 14 DoF, \textbf{realistic} & MuJoCo & \cbmark & \includesvg[height=10pt]{emoji/stands} & \includesvg[height=10pt]{emoji/walk} & \xmark & \textbf{None} & 1e6 / 1200 /   \\
        \hline
        \cite{LeggedRobotFallRecovery} & SAC & \textbf{3D}, 23 DoF, \textbf{realistic} & PyBullet & \xmark & \includesvg[height=4pt]{emoji/supine}\;\includesvg[height=5pt]{emoji/pronea} & \includesvg[height=6pt]{emoji/sitground}\,\includesvg[height=7pt]{emoji/crawl}\,\includesvg[height=8pt]{emoji/squat}\,\includesvg[height=10pt]{emoji/stands} & \xmark & Transition poses &  / / 600  \\
        \hline
        \cite{AgileSoccerSkills} & DMPO & \textbf{3D}, 20 DoF, \textbf{realistic} & MuJoCo & \cbmark & \includesvg[height=4pt]{emoji/supine}\;\includesvg[height=5pt]{emoji/pronea} & \includesvg[height=3pt]{emoji/pushuplow}\,\includesvg[height=6pt]{emoji/layback}\,\includesvg[height=8pt]{emoji/squat}\,\includesvg[height=10pt]{emoji/stands}\,\includesvg[height=10pt]{emoji/walk}\,\includesvg[height=9pt]{emoji/dribble} & \xmark & Pre-programmed & 2e8--2e9 / / 1e6   \\
        \hline
        \textbf{Ours} & PPO & \textbf{3D}, \textbf{32} DoF, \textbf{realistic} & PhysX & \xmark & \includesvg[height=4pt]{emoji/supine}\;\includesvg[height=5pt]{emoji/pronea} & \includesvg[height=8pt]{emoji/rollover}\,\includesvg[height=7pt]{emoji/crawl}\,\includesvg[height=7pt]{emoji/getup}\,\includesvg[height=8pt]{emoji/squat}\,\includesvg[height=10pt]{emoji/stands}\,\includesvg[height=10pt]{emoji/walk} & \cbmark & \textbf{None} & 2e9 / 5000 / 50 \\
        \hline
    \end{tabular}
    }
    \caption*{Pose symbols: \includesvg[height=10pt]{emoji/sitstool} Sitting on a chair;\;
\includesvg[height=10pt]{emoji/stands} Standing;\;
\includesvg[height=5pt]{emoji/legf} Leg on the floor;\;
\includesvg[height=8pt]{emoji/leg} Leg standing;\;
\includesvg[height=8pt]{emoji/squat} Squatting;\;
\includesvg[height=4pt]{emoji/supine} Supine position;\;
\includesvg[height=5pt]{emoji/pronea} Prone position;\;
\includesvg[height=7pt]{emoji/getup} Crouching;\;
\includesvg[height=10pt]{emoji/run} Running;\;
\includesvg[height=6pt]{emoji/sitground} Sitting on the floor;\;
\includesvg[height=10pt]{emoji/walk} Walking;\;
\includesvg[height=3pt]{emoji/pushuplow} Push-up;\;
\includesvg[height=6pt]{emoji/layback} Sit-up;\;
\includesvg[height=6pt]{emoji/bend} Standing forward bend;\;
\includesvg[height=9pt]{emoji/dribble} Dribbling;\;
\includesvg[height=6pt]{emoji/leap} Leap;\;
\includesvg[height=10pt]{emoji/cartwheel} Cartwheel;\;
\includesvg[height=5pt]{emoji/duck} Ducking;\;
\includesvg[height=7pt]{emoji/crawl} Crawling;\;
\includesvg[height=8pt]{emoji/rollover} Rolling over.}
    \caption{Comparison of different algorithms for humanoid robot motion learning.}
    \label{tab:comparison}
\end{table*}

\section{Our Method}

We propose a method that removes traditional constraints, enabling a realistic humanoid robot to learn to stand up from lying down, using only proprioceptive sensor data and without the need for human demonstration or predefined trajectory. Our method is distinct due to the following contributions:
\begin{enumerate}
\item introduced an achievement-triggered multi-path reward function that eliminates the need for human demonstration data or reference motion trajectory; 
\item incorporated an ego-centric representation of state observations, improving data efficiency;
\item effectively learned rhythmic motion by leveraging sine waves with coprime frequencies as central pattern generator signals; 
\item presented an efficient learning strategy under multi-contact constraints by clamping the action output of the model during initial training stages.
\end{enumerate}
These contributions illustrate the developmental principles that can be embedded into the standard reinforcement learning paradigm, as they alter only the state observations, actions, and rewards, with no requirement to modify underlying learning algorithms. We validate the effectiveness of our approach using a continuous-space Advantage Actor Critic algorithm\cite{A2C}, implemented in RL Games\footnote{\url{https://github.com/Denys88/rl\_games}}. This leads to reduced learning resources and an expanded array of learnable tasks without the need to combat catastrophic forgetting. We have tested this approach on a realistic 32-DoF humanoid robot model within a simulation environment, with the robot initially positioned lying on the ground with randomized joint angles.

\subsection{Achievement-triggered multi-path reward function}

Traditional reinforcement learning reward function design concentrates on optimizing a single task through time-invariant linear combinations of positive rewards and negative costs. Recent research has explored reward shaping techniques, such as preference terms\cite{RewardShaping}, or incorporating human preferences into decision trees\cite{TreeReward}, but these may not be suitable for open-ended developmental learning tasks.

We counter this by structuring multiple single rewards as a graph, with each reward representing a motor skill and associated with an achievement score. This score acts as a multiplier to the next reward function node, enabling the robot to progress to more complex motor skills.
Formally, the reward structure is defined as a graph $G = (R, PS)$, where $R$ is the set of reward functions associated with each motor skill. The set $PS$ denotes the condition for activating the next-stage reward function $R_j$ when the current-stage reward $R_i$ reaches a passing score $PS_{ij}$.

At each timestep $t$, the achievement score $A_{ij,t}$ is updated:
$$A_{ij,0} = 0, \;\; A_{ij,t} \leftarrow \max(A_{ij,t-1}, R_{i,t} - PS_{ij})$$

\begin{figure}[h]
\centering
\includegraphics{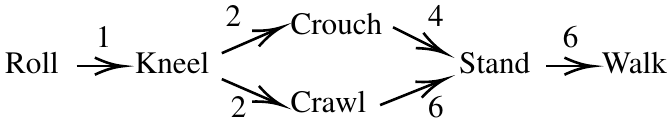}
\caption{Graph-structured reward; the number associated with each edge denotes the passing score}
\label{graph}
\end{figure}

The overall reward $R_t$ is calculated as the sum of the product of each achievement score and the corresponding reward value of the subsequent stage:
$$R_t = \sum_i \sum_j A_{ij,t} R_{j,t}$$
This structure allows more flexible and adaptable modeling of motor skill learning, activating higher-level motor skills only when lower-level ones have been mastered sufficiently. For our motor skill learning problem, the reward is represented as a directed acyclic graph, as shown in Fig.~\ref{graph}. Within this structure, each non-final motor skill reward points to the next through a `passing score' marked on the connecting arrow. Once a reward reaches this passing score, the reward pointed to by the arrow is activated.

The motor skills we focus on include:  
\begin{itemize}
\item \textbf{Rolling over:} Roll from a supine to a prone position.
\item \textbf{Kneeling:} Move from a prone to a kneeling position.
\item \textbf{Crouching:} Achieve an upright torso with legs brought forward and close.
\item \textbf{Crawling:} Propel forward on arms and legs.
\item \textbf{Standing:} Get up from the ground using arms and legs.
\item \textbf{Walking:} Move forward on two legs.
\end{itemize}

\subsection{Ego-centric representation}

In line with developmental theory suggesting early motor skill development relies on an ego-centric coordinate system\cite{egocentric}, we process the simulator's state observations before input into the agent's observation vector. We center all raw poses at the robot's origin and transform all rigid body velocities to relative velocities with respect to the robot's root reference frame. Additionally, sensor measurements are conducted within their respective local frames.

This approach differs significantly from typical reinforcement learning settings, where the simulator's rigid body states in the global frame are used directly as input to the learning agent. Utilizing measurements in the global frame introduces unnecessary variance to the data samples, such as coordinate offsets and widespread heading angles during exploration. This variance reduces sample efficiency as similar robot poses can differ significantly in the global reference frame when translation and rotation are present.

By centering the coordinates, rotations, and velocities to the root reference frame of each robot, we ensure that similar robot poses are represented similarly in the observation space. This facilitates more efficient learning by reducing the variability in the data samples and allowing the agent to better generalize from its experiences.

\subsection{Coprime sine wave CPG signal}

We incorporate central pattern generator (CPG) signals crucial to motion learning by introducing eight pairs of sine waves with coprime frequencies to the agent's state observation vector, defined as:
$$f_{CPG} = [0.5, 0.75, 1.25, 1.75, 2.75, 3.25, 4.25, 4.75]$$
$$CPG_0 = \sin(2 \pi t f_{CPG}), \;\; CPG_\pi = \sin(2 \pi t f_{CPG} - \pi)$$

CPG signals impose strong priors to rhythmic motion generation. To learn rhythms of different speeds, it is necessary to use multiple CPG signals of different frequencies. The aim is to maximize the variety of combinations, or expressive power, of the rhythmic patterns with the least number of CPG signals. To balance this need for expressivity with the physiological constraints of typical human motion, which predominantly falls in the range of 0.5--5 Hz, we selected the first eight prime numbers, dividing them by 4 to map them within this frequency range. To further promote alternate movement in the left and right halves of the body, a phase-inverted copy of the coprime series is added to the observation. This approach allows for the learning of diverse rhythmic patterns while maintaining physiological plausibility and efficiency in the number of signals used.

\subsection{Action clamping}

To simulate the growth of physical strength and provide a safe exploration environment, we introduce a novel approach of action clamping proportionate to the episode progress. The action-torque mapping of the robot joints, based on the stall torques of the actuators, is constrained by a common coefficient that grows linearly from 0 to 1 during the first half of each episode. This means that the robot's actual joint torque limit is only reached midway through the episode.

This method of action clamping is necessary in the early stages of learning, particularly when the robot is lying on the ground with many points of contact. If the actions are not limited, an untrained model that outputs a large action could cause the robot to move violently, hit the ground hard, and generate out-of-distribution force sensor measurements for the learning agent.

By utilizing action clamping, the agent can safely explore the consequences of its joint actions on body poses, enabling the bootstrapping of the policy network's learning. This approach effectively limits the learning agent's achievable states to a smaller range during the early learning stages, thereby improving both training efficiency and overall performance.

\section{Experimental Setup}

This section outlines the experimental setup, including the robot model utilized, the simulator and hyperparameters selected, and the design of the experiments.

\subsection{Robot Model}

The selection of the robot model plays a crucial role in determining the complexity of the learning objectives. Traditional reinforcement learning benchmarks often employ simplified robot models, such as the MuJoCo humanoid\cite{mujoco}. However, these simplifications can veil the challenges encountered when transitioning to more realistic task scenarios. Therefore, to ensure a successful transition from simulation to real-world application, the use of a comprehensive model of a real humanoid robot is crucial.

In light of this, we have chosen the iCub robot\cite{iCub} as our embodied learning agent. This selection was driven not only by the availability of its official URDF model\footnote{https://github.com/robotology/icub-models} and its child-like morphology, aptly designed for developmental robotics research, but also due to its representative and typical configuration in terms of limb geometry and degrees of freedom. Utilizing the latest V3 model, which features 48 rigid bodies and 32 degrees of freedom, we anticipate that our approach should be broadly applicable across a wide range of humanoid robots with similar structures.

Existing methods, efficient on simplified robots such as the MuJoCo humanoid, often fail to directly apply to more realistic robots like the iCub. This underscores the inherent difficulty and need for methods compatible with more realistic robot models. We address this gap through our method, demonstrating that even with these challenging-to-train realistic models, high learning rates and success rates can be achieved. This highlights the versatility and efficacy of our approach, reinforcing the need for methods that are compatible with more realistic robot models.

\subsection{Simulator and Hyper-parameter}

We conducted training in Isaac Gym\cite{IsaacGym} using the PhysX engine, which allows parallel training on a single GPU using vectorized environments. The neural network used by the a2c\_continuous agent is a 4-layer MLP with 800, 400, 200, and 100 units using ELU activation. We used a minibatch size of 65536, a learning rate of 5e-3, and a discount factor of 0.99. Each epoch consists of 32 timesteps, and each episode comprises 3000 timesteps.

\subsection{Experimental Design}

We applied our proposed achievement-triggered multi-path reward function to train the iCub robot to perform a range of gross motor skills. Each motor skill task's individual rewards are defined using torso/limb orientation and root velocity. An episode is terminated when the agent's activity ceases, signaled by all body parts becoming stagnant, suggesting that the output of the agent network has become inactive.

\begin{figure*}[!htb]
     \centering
     \smallskip
     \smallskip
     \begin{subfigure}[b]{0.19\textwidth}
         \centering
         \includegraphics[width=\textwidth]{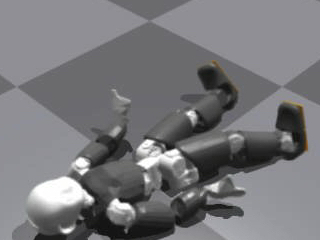}
         \caption{Initial position}
        \label{initial}
     \end{subfigure}
     \begin{subfigure}[b]{0.19\textwidth}
         \centering
         \includegraphics[width=\textwidth]{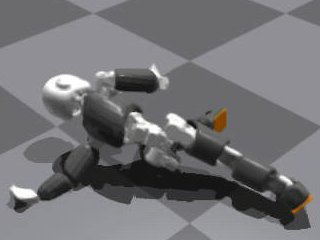}
         \caption{Rolling over}
        \label{roll}
     \end{subfigure}
     \begin{subfigure}[b]{0.154\textwidth}
         \centering
         \includegraphics[width=\textwidth]{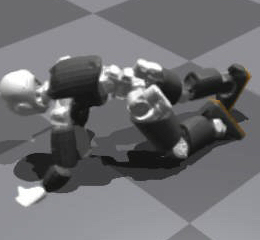}
         \caption{Kneeling}
        \label{kneel}
     \end{subfigure}
     \begin{subfigure}[b]{0.13\textwidth}
         \centering
         \includegraphics[width=\textwidth]{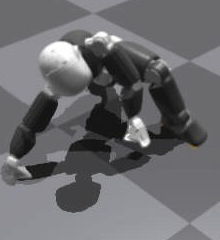}
         \caption{Crouching}
        \label{crouch}
     \end{subfigure}
     \begin{subfigure}[b]{0.107\textwidth}
         \centering
         \includegraphics[width=\textwidth]{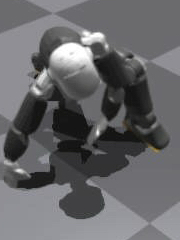}
         \caption{Getting up}
        \label{getting_up}
     \end{subfigure}
     \begin{subfigure}[b]{0.09\textwidth}
         \centering
         \includegraphics[width=\textwidth]{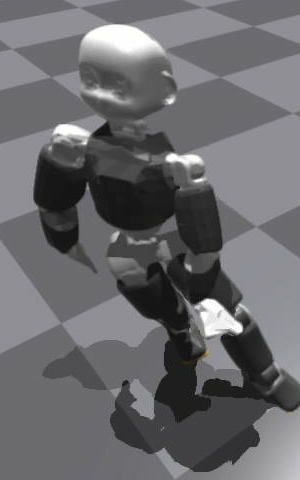}
         \caption{Standing}
        \label{stand}
     \end{subfigure}
     \begin{subfigure}[b]{0.09\textwidth}
         \centering
         \includegraphics[width=\textwidth]{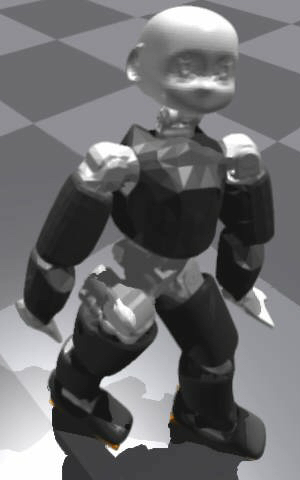}
         \caption{Walking}
        \label{walk}
     \end{subfigure}
        \caption{Learned actions}
        \label{learned_actions}
\end{figure*}

\section{Results}

The successful acquisition of all motor skills outlined in the reward stages graph is demonstrated in Fig.~\ref{learned_actions}. The progression of learned rewards during training reveals a clear pattern: simpler tasks, defined early in the reward structure, are mastered first. These are followed by more complex tasks, culminating in walking, which builds upon all preceding skills. Once the desired skill levels are reached, the rewards remain stable, preventing catastrophic forgetting of previously learned fundamental tasks.

\begin{figure}[!htb]
	\captionsetup[subfigure]{justification=centering}
     \centering
	 \begin{subfigure}[b]{0.235\textwidth}
     \vskip 0pt
         \centering
         \caption{Maximum Reward Over \\ All Episodes}     
         \includegraphics[width=\textwidth]{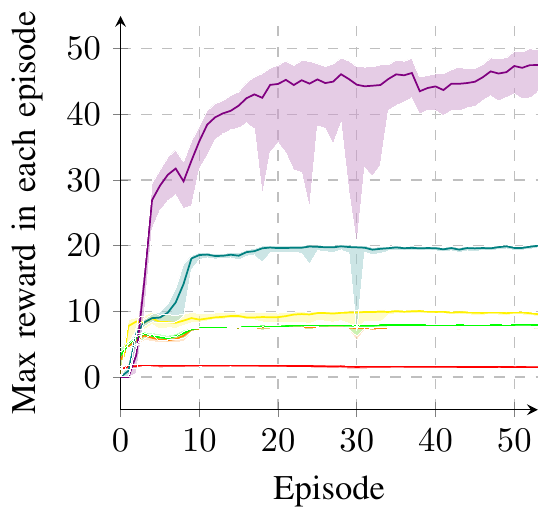}
         \label{max_rew_eps_all}
     \end{subfigure}
     \begin{subfigure}[b]{0.235\textwidth}
     \vskip 0pt
         \caption{Reward Progression Within \\ Episode 53}     
         \centering
         \includegraphics[width=\textwidth]{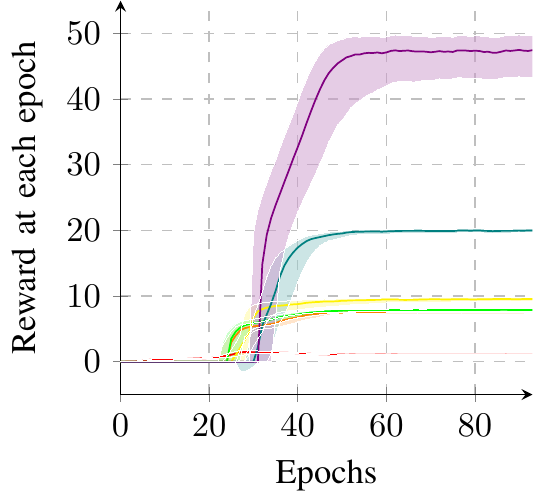}
         \label{best_rew_epo_all}
     \end{subfigure}
	 \begin{subfigure}[b]{0.235\textwidth}
     \vskip 0pt
         \centering
         \caption{Maximum Reward \\ Episodes 0--10}     
         \includegraphics[width=\textwidth]{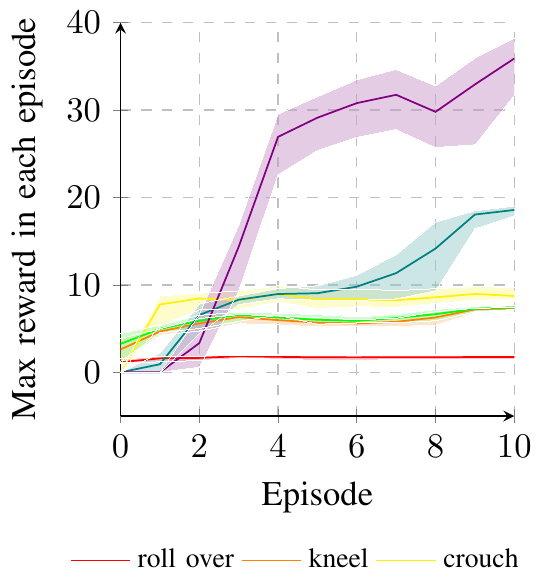}
         \label{max_rew_eps_all_0-10}
     \end{subfigure}
     \begin{subfigure}[b]{0.235\textwidth}
     \vskip 0pt
         \caption{Episode 53 Magnified \\ Epochs 22--38}     
         \centering
         \includegraphics[width=\textwidth]{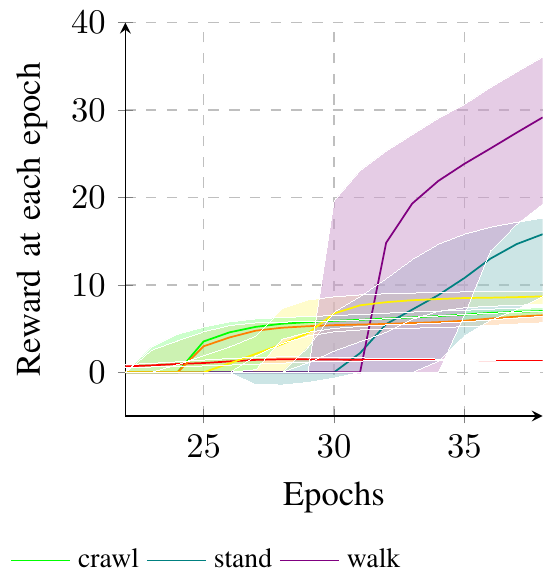}
         \label{best_rew_epo_all_22-38}
     \end{subfigure}
        \caption{Training with ego-centric representation, CPG signal, and action clamping. Shaded colors are plotted between the first and third quartiles. \protect\subref{max_rew_eps_all} and \protect\subref{max_rew_eps_all_0-10} represent the maximum reward achieved for each motor skill over 53 episodes and the first 10 episodes, respectively. \protect\subref{best_rew_epo_all} and \protect\subref{best_rew_epo_all_22-38} depict the reward at each epoch in episode 53 and a detailed view of the stand-up procedure within this episode, respectively.}
        \label{training_all}
\end{figure}

Fig.~\ref{max_rew_eps_all} illustrates the progression of learned rewards throughout the entirety of the training episodes. Each line corresponds to the reward for a specific motor skill. Simpler skills quickly reach their global maximum and remain stable, while more complex tasks, like walking, only contribute to the reward once the preceding task (standing) achieves its passing score. This pattern is depicted more clearly in the zoomed-in view of the first 10 episodes in Fig.~\ref{max_rew_eps_all_0-10}, showing the early acquisition of skills up to standing. Fig.~\ref{best_rew_epo_all} presents the reward curve during episode 53, which achieved the highest overall reward. Within this episode, the robot begins lying on the ground with clamped action and gradually progresses through rolling over, kneeling, crouching, and standing before walking. A more detailed view of the standing procedure within this episode is provided in Fig.~\ref{best_rew_epo_all_22-38}, showing the activation of more advanced skills only after preceding skills' rewards reach their passing scores.

\subsection{Comparison with State-of-the-Art Techniques}

To our knowledge, our proposed methodology, which employs an innovative achievement-based multi-path reward function, is the first single-stage learning setup that enables motion learning on a realistic humanoid model without the need for motion data or a target trajectory. In contrast to the original versions of the PPO and GAE algorithms, which are only successful with unrealistic robot models, our technique enables learning on a realistic humanoid robot with precise geometric constraints and actuator specifications. This significantly simplifies the sim-to-real transfer to the actual iCub humanoid robot or any other existing robot platforms.

Our technique stands out as it eliminates the need for external motion data or pre-programmed trajectory when compared to GAIL, DMPO, and evolutionary methods. Unlike other studies that model pose transition as a graph, our method does not require the robot to initialize in any intermediate poses. Our agent can independently learn to achieve all intermediate and goal poses, starting from a supine position and relying solely on proprioception input.

Our intra-episode skill graph is another distinction from other curricular learning algorithms that learn different skills in separate episodes. This method avoids catastrophic forgetting, as the agent must master the initial skills at the start of every episode to earn rewards on more advanced skills. The multiplicative achievement score further incentivizes the learning agent to refine the proficiency of preceding skills to achieve higher total rewards.

Our method also demonstrates efficiency in terms of training data and computational resources. The learning agent is restricted to task rewards based on its current ability, enabling it to learn to roll over in only two episodes, kneel in four, crouch and crawl in five, stand in ten, and refine its walking skill during episodes 15 to 53. The entire learning procedure requires only 100 minutes on an RTX 4090 GPU.

\subsection{Ablations}

To better understand the contributions of our method's components, we conducted an ablation study. We evaluated the performance of our graph-based reward structure against simpler rewards and structures, and tested the necessity of the developmental learning principles of ego-centric reference frames, CPG signals, and action clamping.

\subsubsection{Single Reward}

A single reward function alone proved insufficient for guiding a high-DoF robot to achieve distant goal states under realistic constraints. The agent exploited the reward function, which resulted in it getting stuck at a local maximum and ultimately ending up in a seated position. Simply summing all types of rewards also resulted in exploitation, with the agent learning to sit and swing its legs to slide forward. These results underscore the limitations of a single reward function for learning diverse motor skills.

\subsubsection{Linear List Reward}

\begin{figure}[h]
  \centering
 \includegraphics{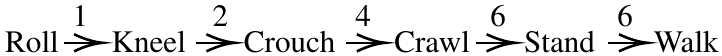}
  \caption{Linear-structured reward; each edge's associated number denotes the passing score}
  \label{linear}
 \end{figure}

A linear list reward structure chains multiple motor skills into a curriculum. However, an inappropriate early stage can impede progress to later stages. When tested with a reward chain of rolling over, kneeling, crouching, crawling, standing, and walking, as shown in Fig.~\ref{linear}, the robot only learned a gorilla-like motion, indicating the need for a more sophisticated reward structure.

\subsubsection{Tree Reward}

\begin{figure}[h]
 \centering
 \includegraphics{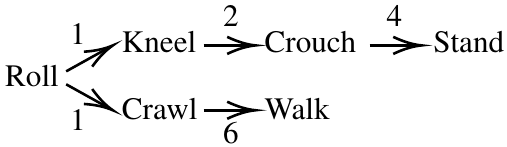}
\caption{Tree-structured reward; each edge's associated number denotes the passing score}
 \label{tree}
\end{figure}

A tree-structured reward scenario offers multiple paths to different end goals, but requires predetermined path design. This rigid structure can result in inactive branches and hinder the learning agent from acquiring intended skills. When tested with a tree reward, as shown in Fig.~\ref{tree}, the robot remained stuck at the prone pose, suggesting the rigidity of a tree reward may not be suitable for open-ended developmental learning tasks.

\subsubsection{Ego-centric reference frame}

\begin{figure}[h]
	\captionsetup[subfigure]{justification=centering}
     \centering
	 \begin{subfigure}[b]{0.235\textwidth}
     \vskip 0pt
         \centering
         \caption{Maximum Reward Over \\ All Episodes}     
         \includegraphics[width=\textwidth]{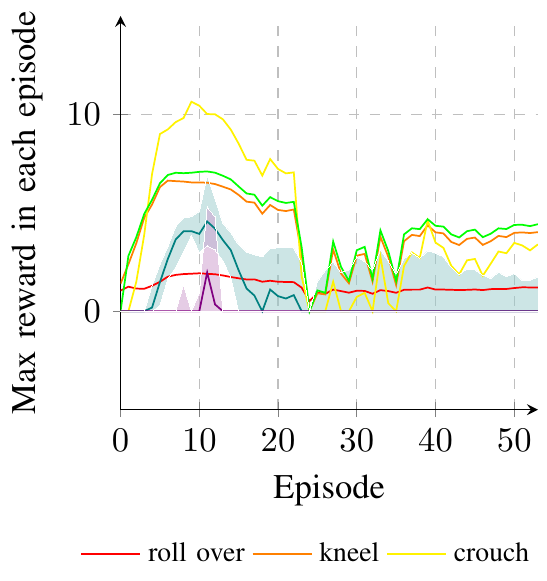}
         \label{max_rew_eps_no_ego}
     \end{subfigure}
     \begin{subfigure}[b]{0.235\textwidth}
     \vskip 0pt
         \caption{Reward Progression Within \\ Episode 11}     
         \centering
         \includegraphics[width=\textwidth]{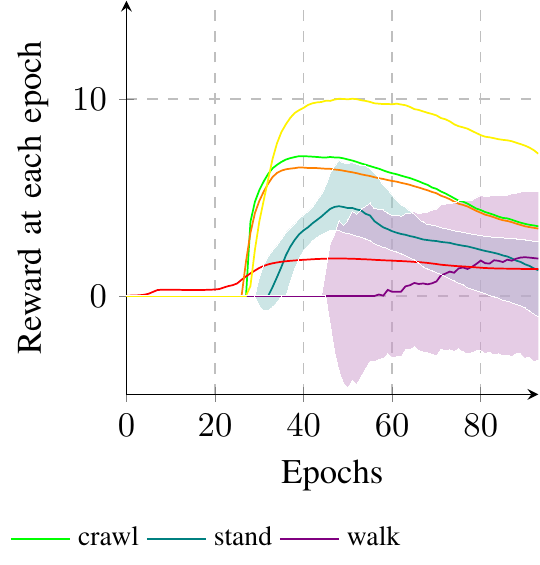}
         \label{best_rew_epo_no_ego2}
     \end{subfigure}
        \caption{Training with CPG signal and action clamping, without ego-centric representation. The quartile ranges for standing and walking are plotted with color shades for clarity.}
        \label{training_no_ego}
\end{figure}

Relying on the world reference frame for pose and velocities significantly hampers the learning of motor skills. Most agents had difficulty rolling over and subsequently transitioned into a kneeling position. This reduced sample efficiency suggests the agent's ability to generalize and effectively explore the parameter space is limited without an ego-centric representation. This is illustrated in Fig.~\ref{training_no_ego}, which shows the impact of excluding the ego-centric representation on the learning process.

\subsubsection{CPG signals}

\begin{figure}[h]
	\captionsetup[subfigure]{justification=centering}
     \centering
	 \begin{subfigure}[b]{0.235\textwidth}
     \vskip 0pt
         \centering
         \caption{Maximum Reward Over \\ All Episodes}     
         \includegraphics[width=\textwidth]{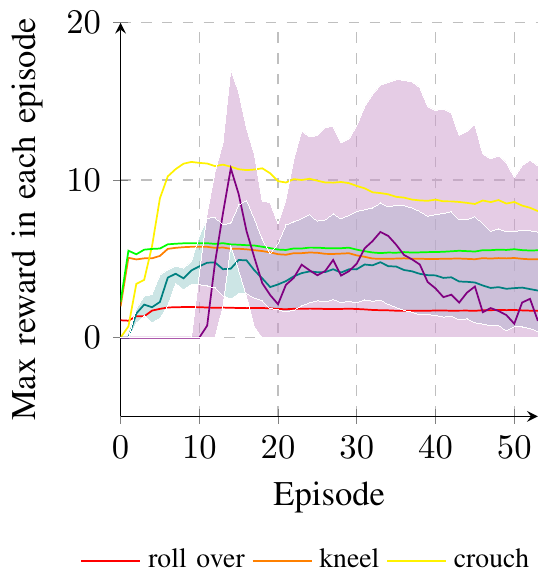}
         \label{max_rew_eps_no_cpg}
     \end{subfigure}
     \begin{subfigure}[b]{0.235\textwidth}
     \vskip 0pt
         \caption{Reward Progression Within \\ Episode 14}     
         \centering
         \includegraphics[width=\textwidth]{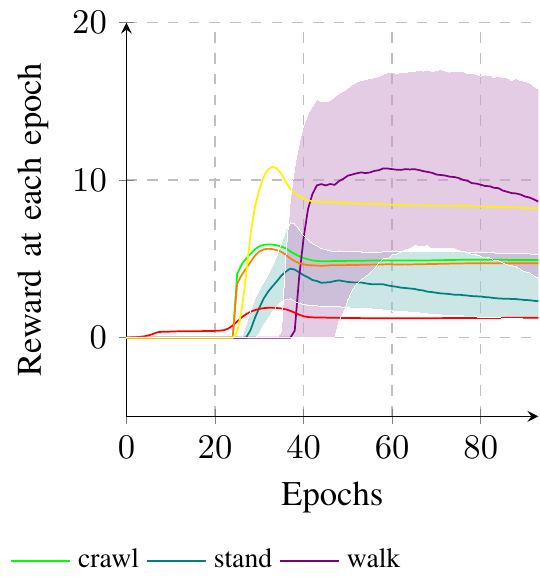}
         \label{best_rew_epo_no_cpg2}
     \end{subfigure}
        \caption{Training with ego-centric representation and action clamping, without CPG signal.}
        \label{training_no_cpg}
\end{figure}

The absence of the CPG signal during training prevents the robot from acquiring motor skills beyond kneeling and crouching. Fig.~\ref{training_no_cpg} shows the diminished progress in skill acquisition without the CPG signal.

In a different scenario, when the CPG signal was removed only during deployment, the robot exhibited disorientation and instability when transitioning from standing to walking, along with a reduced step size. This behavior, illustrated in Fig.~\ref{no_cpg_at_testing_sf}, underscores the importance of the CPG signal in acquiring advanced motor skills and ensuring smooth and stable motor control during deployment.

\subsubsection{Action clamping}

\begin{figure}[h]
	\captionsetup[subfigure]{justification=centering}
     \centering
	 \begin{subfigure}[b]{0.235\textwidth}
     \vskip 0pt
         \centering
         \caption{Maximum Reward Over \\ All Episodes}     
         \includegraphics[width=\textwidth]{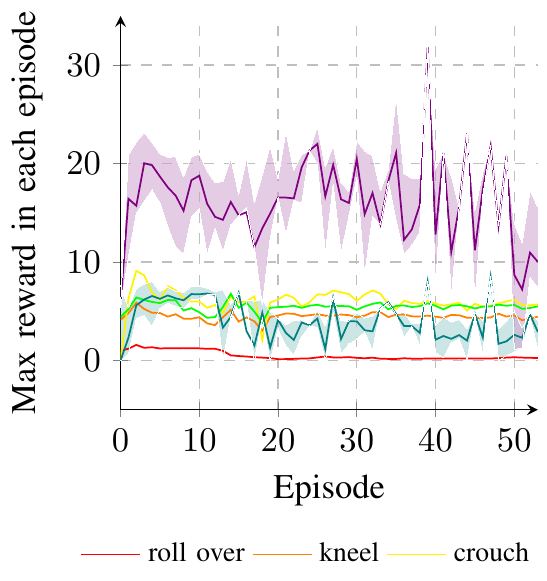}
         \label{max_rew_eps_no_clamp}
     \end{subfigure}
     \begin{subfigure}[b]{0.235\textwidth}
     \vskip 0pt
         \caption{Reward Progression Within \\ Episode 3}     
         \centering
         \includegraphics[width=\textwidth]{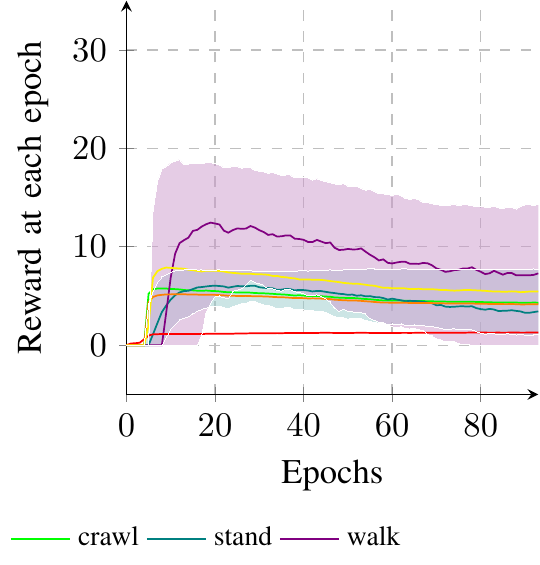}
         \label{best_rew_epo_no_clamp}
     \end{subfigure}
        \caption{Training with ego-centric representation and CPG signal, without action clamping.}
        \label{training_no_clamping}
\end{figure}

Action clamping proved to be a vital component for the successful learning of complex motor skills. Without a progressive increase in the actuator joint torque limit, most learning agents were unable to complete the task of rolling over, indicating the essential role of action clamping in learning complex motor skills. This is showcased in Fig.~\ref{training_no_clamping}, which details the limited progress in skill acquisition without action clamping.

\begin{figure}[h]
	\captionsetup[subfigure]{justification=centering}
     \centering
     \begin{subfigure}[b]{0.235\textwidth}
     \vskip 0pt
         \centering
         \caption{Without CPG signal}
         \includegraphics[width=\textwidth]{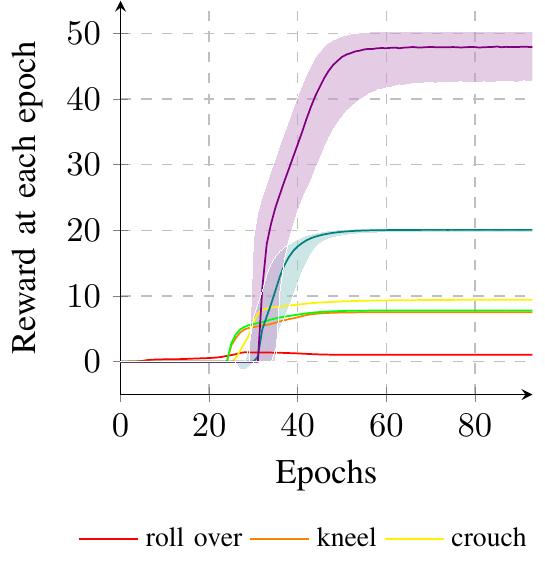}
         \label{no_cpg_at_testing_sf}
     \end{subfigure}
     \begin{subfigure}[b]{0.235\textwidth}
     \vskip 0pt
         \centering
         \caption{Without action clamping}
         \includegraphics[width=\textwidth]{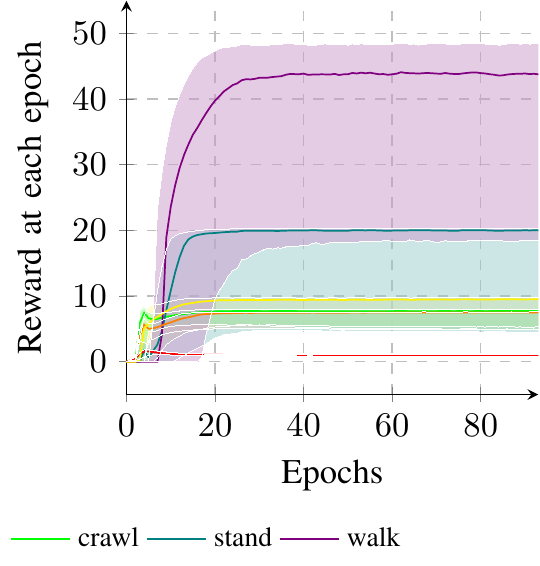}
         \label{full_power_at_testing_sf}
     \end{subfigure}
        \caption{Testing on successfully trained model.}
        \label{all_testing}
\end{figure}

When action clamping was removed only during deployment, the robot's motion was less stable and the maximum reward obtained was reduced. Despite achieving the goal state more quickly, the robot reached only 90\% of the original reward for walking and 83\% for standing, with significantly lower first quartile values, as shown in Fig.~\ref{full_power_at_testing_sf}.

\addtolength{\textheight}{0cm}   


\subsection{Discussion}

The findings from this study highlight the effectiveness of our achievement-triggered multi-path reward design in enabling the iCub robot to acquire a wide array of gross motor skills. However, to make this approach more applicable to real-world scenarios, future work will need to integrate goal-conditioned rewards during the training stage. This will enable dynamic specification of desired behaviors at deployment, allowing the robot to adapt to evolving task requirements and perform more complex actions with enhanced precision.

A video supplement provides a visual demonstration of the real-time acquisition of the target motor skills for multiple robots in parallel and the results of the ablation study.

\section{CONCLUSIONS}

In conclusion, this study presents a novel methodology for enabling humanoid robots to learn a diversity of motor skills based on principles derived from developmental robotics. Our achievement-triggered multi-path reward function allows the robot to learn gross motor skills without the need for explicit programming or human demonstrations. The results from our experiments demonstrate that our approach outperforms traditional methods in terms of success rates and learning speed, underscoring the potential of developmental reinforcement learning in the field of robotics. 

Importantly, the sim-to-real transfer potential of our method is enhanced by our use of a realistic model of the iCub robot in simulations, which has accurate geometry and physics modeling. Our action clamping setup promotes the safety of the robot during real-world fine-tuning, reducing the risk of mechanical damage. Looking forward, our method could be extended to fine motor skills learning, enabling robots to master an even wider array of skills for performing tasks in real-world scenarios.

\bibliographystyle{IEEEtran}
\bibliography{ref}

\end{document}